\documentclass{article}

\newcommand*\samethanks[1][\value{footnote}]{\footnotemark[#1]}
\usepackage{graphicx,dcolumn,hyperref,xcolor,bm,natbib}
\usepackage{calligra}
\usepackage{egothic}
\usepackage[T1]{fontenc}
\usepackage{fontawesome5}
\usepackage{amsmath}
\usepackage{epsfig}
\usepackage{units}
\usepackage[utf8]{inputenc}
\usepackage[preprint]{neurips_2023}
\usepackage{multirow}

\usepackage[utf8]{inputenc} 
\usepackage[T1]{fontenc}    
\usepackage{hyperref}       
\usepackage{url}            
\usepackage{booktabs}       
\usepackage{amsfonts}       
\usepackage{nicefrac}       
\usepackage{microtype}      
\usepackage{xcolor}         

         \newcommand{\cM}{{\cal M}}       
\newcommand{\pL}{\left(} \newcommand{\pR}{\right)}

\title{WavPool: A New Block for Deep Neural Networks}

\author{%
  Samuel D.~McDermott\thanks{Equal contribution.} \\
  Department of Astronomy and Astrophysics\\
  University of Chicago, Chicago, IL \\
  \texttt{samueldmcdermott@gmail.com}
  \And
M.~Voetberg\samethanks \\
  Fermi National Accelerator Laboratory \\
  Scientific Computing Division, Batavia, IL \\
  \texttt{maggiev@fnal.gov}
  \AND
Brian Nord \\
  Fermi National Accelerator Laboratory, Scientific Computing Division, Batavia, IL \\
  Department of Astronomy and Astrophysics, University of Chicago, Chicago, IL \\
  Kavli Institute for Cosmological Physics, University of Chicago, Chicago, IL \\
  Laboratory for Nuclear Science, MIT, Cambridge, MA \\
  \texttt{nord@fnal.gov}
  }

\begin{document}

\maketitle

\begin{abstract}
    Modern deep neural networks comprise many operational layers, such as dense or convolutional layers, which are often collected into blocks. 
    In this work, we introduce a new, wavelet-transform-based network architecture that we call the {\it multi-resolution perceptron}: by adding a pooling layer, we create a new network block, the {\it WavPool}.
    The first step of the multi-resolution perceptron is transforming the data into its multi-resolution decomposition form by convolving the input data with filters of fixed coefficients but increasing size. 
    Following image processing techniques, we are able to make scale and spatial information simultaneously accessible to the network without increasing the size of the data vector.
    WavPool outperforms a similar multilayer perceptron while using fewer parameters, and outperforms a comparable convolutional neural network by $\sim 10\%$ on relative accuracy on CIFAR-10.
    \footnote{{\tt FERMILAB-CONF-23-278-CSAID}}
\end{abstract}

\section{Introduction}

Discrete signals have the potential to represent an amount of information that scales exponentially with the number of bits.
Despite this fact, 
modern computational methods have demonstrated an ability to abstract patterns from data with far less than an exponential degree of complexity.
Sometimes, the only independent variable underlying these patterns is their location -- e.g., position in a sentence provides nearly all of the information necessary for natural language processing.
In other signals, patterns are present that are not apparent as a function of the given variables but are present instead in the variables of a conjugate space.
Frequency and time are conjugate to one another, which is why Fourier analysis is useful in analyzing sounds.
Spatial scale and position are conjugate to one another, which is why spherical harmonic analysis is useful in analyzing the cosmic microwave background \citep{1970ApJ...162..815P, 1984ApJ...285L..45B}. 
Most data, however, are encoded in signals that are not organized in any single variable, instead exhibiting regularities of both position and scale that are not simply accessible from spatial or scale data alone.
This suggests a need for a type of signal processing that remains simultaneously sensitive to conjugate pairs of variables.

Many methods have been used to extract information and discover patterns in challengingly complex data sets.
For example, matched filters have a long and successful history in the field of signal processing \cite{1057571}. 
Recently, deep learning has upended long-held notions of what is possible in pattern recognition, with significant implications for science and society. 
For example, the multi-layer perceptron (MLP) \cite{rosenblatt1958perceptron} gave the first hint of how stacking many layers of matrix arithmetic between layers of nonlinearity could produce adaptable and flexible learning systems \cite{Cybenko1989ApproximationBS, Hornik1989MultilayerFN}. 
Architectures that further generalized these techniques, like convolutional neural networks (CNNs) \cite{cnnLeCun1998} and the transformer \cite{Attention2017arXiv170603762V}, have proven transformative in extending deep learning techniques to two-dimensional images and text, respectively. 
Nevertheless, traditional architectures require significant complexity (e.g., multiple stacked convolutional layers) to identify data features.

Wavelets are mathematical tools that can be used to provide simultaneous sensitivity to conjugate data -- e.g., position and size or time and frequency.
Therefore, they have the potential to provide unique insights on data that are not purely characterized by spatial or scale information. 
Wavelet families comprise functions of finite spatial extent that are related to one another through scaling and translational transformations.
If a family provides a complete, orthonormal basis of functions, then convolving the family with a set of data will provide a decomposition of the data. 
Wavelets were first discovered in the early 20th century, but many of their more remarkable properties were not formalized, generalized, or fully explored until the 1980s and 1990s \citep{daubechiesbook, 192463, Meyerreview, 488696}.
Methods based on the wavelet transform have since played a foundational role in 
many fields.
For example, they undergird modern computational methods like the JPEG image compression algorithm \citep{2001ISPM...18...36S}. 

The wavelet transform and related functions like the Laplacian pyramid technique have been used in combination with deep learning techniques to achieve state-of-the-art performance on various image-related tasks -- image reconstruction with autoencoders \citep{2017arXiv171207493C} or more generally convolutional layers \citep{2017arXiv170403915L, DBLP:journals/corr/abs-1805-07071}, image classification \citep{DBLP:journals/corr/abs-1805-08620}, representation learning \citep{2022arXiv220704978Y}, and learnable wavelets for denoising with reinforcement learning \cite{2022PNAS..11906598M}.

In this work, we combine wavelets that have compact support \citep{192463} with deep learning operations in a new method. 
We use Daubechies wavelets \citep{daubechiesbook} to decompose data into components on different scales without increasing the size of the data vector.
The wavelet decomposition facilitates training on features of different size and scale in parallel across all length scales spanned by the input.
This is not possible for networks comprised of filters of fixed size, as in a CNN.
By construction, this will contain the same information content as the original, non-decomposed data without increasing the size of the data vector.
However, spatial and scale information about the data will be made simultaneously accessible to the neural network.
We train a series of MLPs in parallel on the decomposed images.
We refer to these multiple, parallel MLPs as a multi-resolution perceptron (MRP).
We use a pooling layer to extract a classifier from the MRP.
This trainable block is called ``WavPool.''

\section{Theory of Multi-Resolution Decomposition} 
\label{sec:theory}

In this section, we provide a review of some basic results in wavelet methods.
The reader who is familiar with this formalism can skip to Sec.~\ref{sec:method}.

Consider an input signal $S$ defined on a D-dimensional grid. 
In this work, we will use a wavelet transform to partition this data into its multi-resolution decomposition (MRD).
The MRD of $S$ is a set of transformed signals,
\begin{equation} \label{eq:decomp}
    \cM_S = \{ C(S), W_L^a(S), \dots , W_1^a(S) \},
\end{equation}
where $a$ is an index that depends on the dimensionality of $S$; $L$ is the number of levels of the decomposition, which depends on the wavelet that is used for the decomposition; $W(S)$ are the {\it details} of $S$, which are the features of $S$ at a particular spatial scale; and  $C(S)$ is the {\it smoothest view} of $S$, which is its global average for the Haar wavelet, but is a non-trivial matrix for other wavelets.
The size of a signal $[S]$ is equal to the sum of the sizes of all of the constituent signals $\zeta$ that comprise its MRD: $[S] = \sum_{\zeta \in \cM_S} [\zeta]$.
For a D-dimensional signal, the index $a$ in Eq.~(\ref{eq:decomp}) takes on $2^{\rm D} - 1$ different values. 
For example, for a 2-dimensional signal, $a$ takes on three values, corresponding to the horizontal, vertical, and diagonal details at each level, as described in more detail presently.
The number of levels $L$ is set by the size of the signal and the {\it number of vanishing moments} $n_v$ of the wavelet used in the decomposition:
\begin{equation} 
\label{eq:num_levels}
    L = \lfloor \log_2 [S] \rfloor - n_v + 1.
\end{equation}
The wavelets used in this work, first described by Daubechies \citep{daubechiesbook}, are indexed by the number of vanishing moments they have, meaning the degree of the polynomial function that they can set to zero: the Daubechies-$n_v$ wavelet will set a polynomial of degree $n_v-1$ to zero.

We now describe the components of the MRD. 
This requires two functions: a {\it smoothing wavelet} $\phi$ and a {\it differencing wavelet} $\psi$.
The differencing wavelet is given by reversing the smoothing wavelet while simultaneously alternating the parity.
If $\phi$ has $N$ entries indexed by $i$, then the $i^{\rm th}$ entry of $\psi$ is $\psi_i = (-1)^i \phi_{N-i}$.
In Eq.~(\ref{eq:decomp}), $C(S)$ is the smoothest view of $S$: this is also the mean of $S$, and it is obtained by
\begin{equation} 
\label{eq:C_av}
    C(S) =  \phi \underbrace{\circ (\phi \circ ( \cdots }_{L\, {\rm times}} S )),
\end{equation}
where $\circ$ is the convolution operation and $L$ is the maximum number of times $\phi$, which is the same as the number of levels (see Eq.~\eqref{eq:num_levels}).
Each of the products in the convolution reduces the size of the input, so that each level of the MRD is a different size.
The convolution procedure is applied iteratively to the smoothed images until a single number $C(S)$ remains. 
One can generalize the smoothest view $C(S)$ from Eq.~(\ref{eq:C_av}) to signals that have been partially smoothed to level $\ell$:
\begin{equation} 
\label{eq:Ci}
    C_\ell(S) =  \phi \underbrace{\circ (\phi \circ ( \cdots }_{\ell\, {\rm times}} S )),
\end{equation}
where $\ell$ is the index of the spatial scale -- the $C_\ell(S)$ for larger values of $\ell$ have been increasingly smoothed.
The $W_\ell^a(S)$ from Eq.~(\ref{eq:decomp}) are the {\it details} of $S$.
They are given by applying a differencing wavelet $\psi^a$ on the partially smoothed images:
\begin{equation} 
\label{eq:Wi}
    W_{\ell+1}^a(S) = \psi^a \circ C_\ell(S) = \psi^a \circ ( \phi \underbrace{\circ (\phi \circ ( \cdots }_{\ell\, {\rm times}} S ))).
\end{equation}
$W^a_\ell(S)$ represents features of $S$ that exist at different spatial scales and exist for different spatial symmetries (as indexed by $a$).
For larger values of $\ell$, $W_\ell^a(S)$ represents features of $S$ that have larger spatial extent.
The essential feature of the MRD is that different levels are of different sizes and contain distinct information about different spatial scales. 
This feature enables partitioning of the data to provide access to different information than is provided by the original data.

In this work, we will exclusively use the Daubechies-1 wavelet ($n_v = 1$), also known as the Haar wavelet.
For the specific case of the one-dimensional Haar wavelet, the smoothing and differencing wavelets are
\begin{equation} \label{eq:Haar1D}
    \phi_{1,1} = \frac1{\sqrt2} \pL \begin{array}{cc} 1 & 1 \end{array} \pR, \qquad \psi_{1,1} = \frac1{\sqrt2} \pL \begin{array}{cc} 1 & -1 \end{array} \pR,
\end{equation}
which are convolved over the input signal $S$ with stride two. 
Therefore, the $W_\ell(S)$ details are half as long as the $C_{\ell-1}(S)$ smooth images from which they are obtained. 
For two-dimensional inputs, the Haar wavelets are
\begin{equation} \label{eq:Haar2D}
\begin{split}
    \phi_{1,2} = \phi_{1,1} \otimes \phi_{1,1}, \qquad \psi_{1,2}^v = \psi_{1,1} \otimes \phi_{1,1},\\ \psi_{1,2}^h = \phi_{1,1} \otimes \psi_{1,1}, \qquad \psi_{1,2}^d = \psi_{1,1} \otimes \psi_{1,1},
\end{split}
\end{equation}
where $\otimes$ is the outer product and $a=v,h,d$ designate the vertical, horizontal, and diagonal differences that are possible in two dimensions. 
For an MRD of a two-dimensional image, the two-dimensional Haar smoothing wavelet $\phi_{1,2}$ is a $2\times2$ matrix, all of whose entries are $1/2$.
The two-dimensional Haar differencing wavelets $\psi_{1,2}^a$ are also $2\times2$ matrices.
Thus, $C_\ell(S)$ and the $W_\ell^a(S)$ each contain one-quarter as many entries as the $C_{\ell-1}(S)$ from which they are obtained.
The Haar wavelet also has the property that the sums of the details at level $\ell$ equal the difference of the smoothed images at levels $\ell-1$ and $\ell$, when the signal at level $\ell$ have expanded by repeating each entry. 
This is reminiscent of the Laplacian pyramid algorithm \citep{1095851}, except that the MRD generates these details constructively and provides orientation information in dimensions larger than one.
We provide worked examples of the 2-dimensional Haar decomposition in Apps.~\ref{sec:App_matrixdecomp} and \ref{sec:App_imagedecomp}.

Importantly, the contents of the MRD, the set $\cM_S$, form a {\it partition} of $S$, such that $S$ can be reconstructed via
\begin{equation} \label{eq:reconstruction}
    S = \phi^{-1} C(S) + \sum_{\ell=1}^L \sum_a (\psi^{-1})^a W_\ell^a(S),
\end{equation}
where $\phi^{-1}$ and $\psi^{-1}$ are the  ``reconstruction'' or ``inverse'' wavelets, respectively. 
The numerical values of these depend on the conventions adopted for the MRD.
We show a concrete example in the appendix where the inverse wavelets are identical to the wavelets used in the MRD.

One implication of Eq.~(\ref{eq:reconstruction}) is that $\cM_S$ and $S$ are two fully equivalent, lossless representations of exactly the same information. 
This can be confirmed by counting degrees of freedom. 
For a one-dimensional input $S$ of length $2^L$ that we have decomposed with the Daubechies-1 wavelet, $W_\ell(S)$ has length $2^{L-\ell}$, and $\cM_S$ has length $1+\sum_{\ell=0}^{L-1} 2^i = 2^L$, where the leading $1$ comes from $C(S)$ and the remaining terms come from the $W_\ell(S)$.
Despite retaining all the information of the original image, the MRD $\cM_S$ has rearranged this data in a novel way according to scale and, in dimensions larger than one, orientation.

The MRD reorganizes the information contained in the input so as to enable direct access to information that is encoded in any way other than purely through local spatial correlations.
Heuristically, the arrangement of an MRD is similar to telescopic series.
This makes the MRD a convenient set of inputs to a DL network if we want that network to learn -- in a size-agnostic way -- aspects of data aside from those encoded purely through spatial locality.
This stands in contrast to a CNN, which encodes spatial information through filters of fixed, predetermined size. 

In this work, we will henceforth only use the Haar wavelet applied to input signals $S$ with two dimensions -- e.g., images.
This fixes the index $a$ to take on three values -- corresponding to vertical, horizontal, and diagonal differences, as in Eq.~\eqref{eq:Haar2D}.
For notational convenience, we will drop the explicit dependence on $S$ in the average and the details. 

We summarize many of the technical terms we use in this work in App.~\ref{sec:App_Terminology}.
We also connect our terminology to alternate terminology found throughout the literature.

\section{New Method: The WavPool Block}
\label{sec:method}

\begin{figure}[t]
    \centering
    \includegraphics[width=0.65\textwidth]{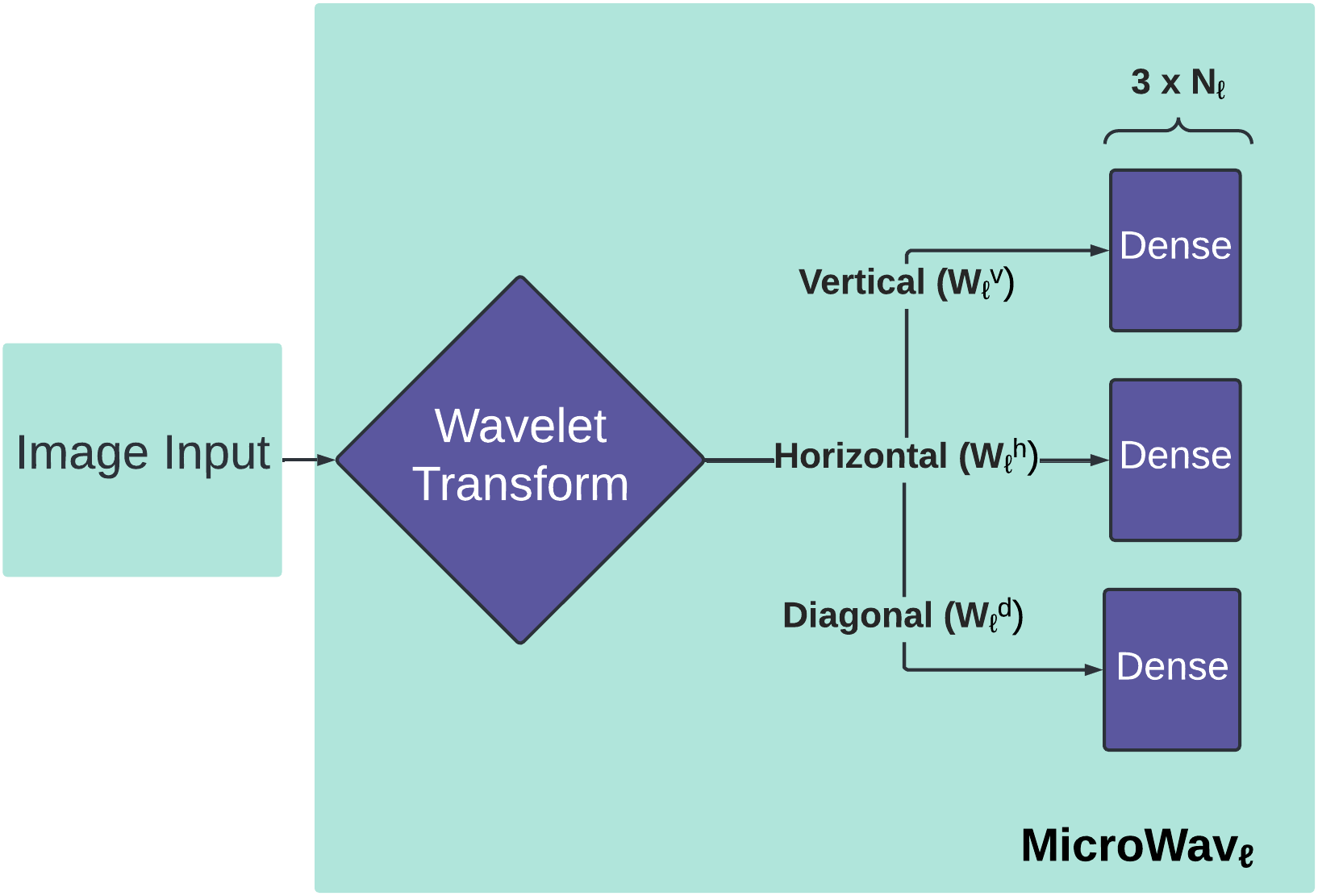}
    \caption{
    The MicroWav operation. 
    The three-tailed operation applies the wavelet decomposition at a defined level $\ell$ and assigns each of the three details to their own dense layer of size $N_\ell$. 
    }
    \label{fig:microwav}
\end{figure}

\begin{figure}[t]
    \centering
    \includegraphics[width=0.95\textwidth]{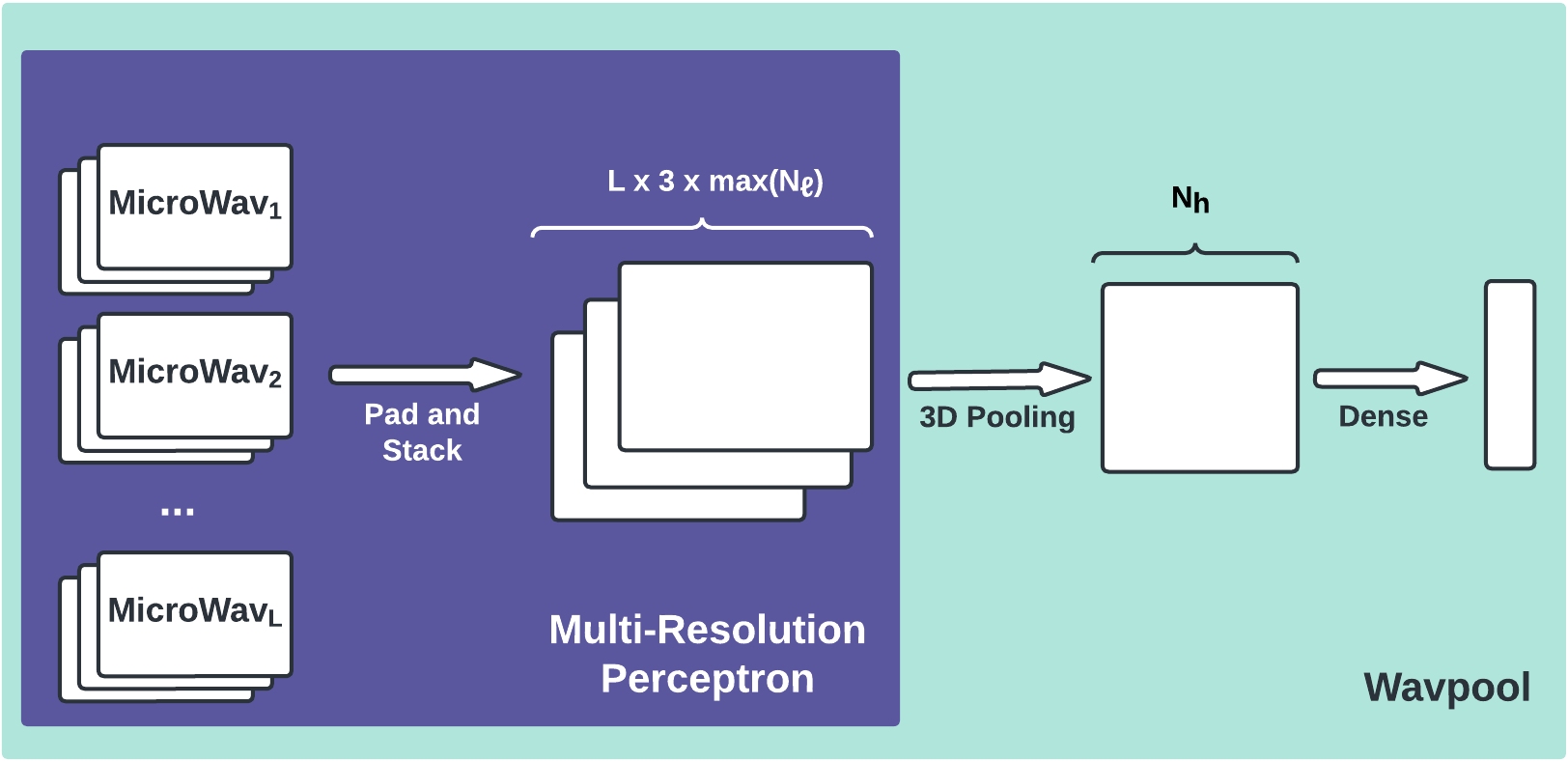}
    \caption{Architectural diagram of the WavPool block. 
    The block comprises individual hidden-layer triplets (``MicroWav''; Fig.~\ref{fig:microwav}) that account for each level of the MRD, which are then fed into a pooling layer.
    This produces an output with shape $ L \times 3 \times \max(N_\ell) $.
    A 3D Pooling operation is applied to this output, followed by a dense layer.  
    The breakdown of parameter and time complexity of each point in the network can be seen in Table~\ref{table:wavpool_complex}
    }
    \label{fig:wavpool_arch}
\end{figure}

We introduce the {\it WavPool} block to incorporate the wavelet decomposition into a neural network context.
The WavPool takes in the levels of the MRD on an input image and applies a dense layer to each component of the transform.
We refer to the output of this collection of many dense layers as the multi-resolution perceptron (MRP).
The {\it Pool} of WavPool is derived from the fact that we apply a maximum pooling operation to the MRP.
The pooling is critical to capture the different layers of granularity of the transform and allow the training process to determine the importance of different decomposition components. 

The fundamental {\it MicroWav} layer is shown in Fig.~\ref{fig:microwav}.
Each MicroWav component, which makes up the first part of the WavPool, operates on a specific level $\ell$ of the MRD. 
Each image detail $W_\ell^a$ is fed into one of three independent dense layers.
The output of each level $\ell$ has its own hidden size, $N_\ell$.
After performing experiments for an optimized network, as discussed in Sec.~\ref{sec:experiments}, we find that hidden layers of different sizes lead to better performance. 

To turn the output of the MicroWav layers into a classifier, we first collect the dense outputs corresponding to all $L$ levels of the MRD.
We call this combined output a multi-resolution perceptron (MRP): it is inspired by the perceptron \cite{rosenblatt1958perceptron}, but the data has been processed by the MRD.
We wish to train on all of the content of the MRP, but we also want to allow the layers to have different sizes to accommodate the different information content of each layer of the MRD.
We fix the hidden layer sizes to scale inversely with $\ell$, such that $\ell \times N_\ell$ is constant.
This has multiple implications.
First, this scaling reduces the number of trainable parameters in the network because higher MicroWav levels have fewer outputs.
Second, because this scaling is sub-exponential, the higher MicroWav levels are smaller.
This helps the network avoid over-fitting to higher levels of the decomposition, which would otherwise have an unbalanced node-to-input feature ratio.
Finally, this ensures that the MRD is not inverted at the end of the network, as it is in previous embeddings of wavelets in neural networks.
Instead, this choice enforces {\it sparsity} inside the WavPool, effectively compressing the lower levels of the decomposition.

To enable a pooling operation on the differently sized outputs of the MicroWav layers, we apply a padding operation to the output of each of these sub-blocks.
This ensures a uniform shape across the block, such that the shape of the stacked output of the MRP is $ L \times 3 \times N_1$.
The resulting block is compressed by a pooling operation.
We name the entire network the {\it WavPool}.
The complete WavPool diagram is shown in Fig.~\ref{fig:wavpool_arch}.

\begin{table}[b]
\caption{The time complexity and resulting matrix size for each step in the WavPool operation. 
$N_\ell$ is the size of the output of the dense network that has been applied to each {\it detail} $W_\ell$; $k_n$ is the selected kernel size, a hyperparameter for the shape of the pooling operation; and $m$ is the selected output size.}
\centering
  \begin{tabular}{lll}
    Step   & Output Size  & Time Complexity \\
    \hline
    MicroWav (single $\ell$) & $ 1 \times 3 \times N_\ell $ & $O(n + 3n^3)$ \\
    Padding/Stacking & $ L \times 3 \times N_1 $ & $O(L)$\\
    3D Pooling & $ (L-k_1+1) \times (4-k_1) \times (N_1-k_2+1) $ & $O(n^3)$\\
    Dense Output & $ m $ & $O(n^2m)$ \\
    \end{tabular}
    \label{table:wavpool_complex}
\end{table}

In Tab.~\ref{table:wavpool_complex}, we estimate the time complexity of components of the WavPool.
The WavPool block itself derives most of its time complexity from the dense layers that compose the majority of its calculations. 
The Haar wavelet itself has linear time complexity with the size of its input $n$, and the operations within MicroWav are sequentially applied. 
Thus, the  MicroWav has time complexity of $O(n)+O(\sum_{t=0}^{T} N_t \times n^2)$.
Within the MicroWav, $T=3$, so this complexity can be simplified to $O(n+3n^3)$, or $O(n^3)$.
This further applies to the full block.
The stacking operation takes place in constant time, and the 3D Pooling operation is an operation $O(n^3)$. 
As all these operations are linearly applied, this produces a block with time complexity of $O(H)+O(\sum_{t=0}^{T} N_t \times H^2)+ O(L) + O(\prod_{i=0}^3(k_i) + O(m \times N_\ell^2)$, or more compactly, $O(n^3)$
The matrix multiplication required to compute dense layers is also $O(n^3)$.
Therefore, the MLP we compare to has a time complexity of $O(2n^3)$, putting the complexity on par with the WavPool. 
Convolutional networks are more dependent on the size of their filters and number of channels, which is described as $O(m_{in} \times m_{out} \times \prod_{i=0}^D k_i^2)$, \cite{DBLP:journals/corr/He014}, where $m$ refers to the channels in the convolution and $k_D$ is the kernel size in each dimension. 
Because $k_i$ has a maximum size of $n$ (i.e., the kernel cannot exceed the size of the input data), in the case of our 2D experiment we constrain the complexity of the CNN to be $O(m_{in}\times m_{out}\times n^4)$ at a maximum.
In some circumstances (e.g., with very large kernels or with many in and out channels), the CNN is more time-intensive than the maximum-complexity form of the WavPool.

\section{Experiments: Comparing WavPool, MLP, and CNN}
\label{sec:experiments}

To investigate the efficacy of WavPool networks, we conduct experiments that compare the WavPool network, a two-hidden layer dense network, and a CNN.
The CNN comprises two convolutional layers, each of which has a batch norm layer between them. 

We test the networks on multiple datasets: MNIST \cite{lecun2010mnist}, Fashion MNIST \cite{xiao2017online}, and CIFAR-10 \cite{Krizhevsky2009LearningML}. 
In our experiments, CIFAR-10 was transform into greyscale to constrain the problem to two-dimensional space.
We perform three trials on each data set with each network.
For each trial, we changed the initialization weights and the subset of each dataset used by setting a different random seed.
For all experiments, we use 4000 training images and 2000 validation images. 
Each trial is run for 120 epochs maximum -- though subject to early stopping if the validation loss stalled after waiting for five epochs of patience. 
Networks are built with PyTorch \cite{DBLP:journals/corr/abs-1912-01703}.
The wavelet-based networks are built with PyWavelet \cite{Lee2019} as a calculation method for the wavelet decomposition. 
All calculations were performed on a 10-core CPU. 

\begin{table}[b]
  \caption{Parameter count and validation score metrics for non-optimized networks. 
  The network hyperparameters were arbitrarily chosen, along with the width of the networks/kernel size.
  The receiver operating characteristic (ROC) curve and accuracy values are for each network are obtained from a validation set. 
  Without tuning, the WavPool performs the best on two out of the three tasks. 
  The MLP narrowly outperforms WavPool on the Fashion-MNIST task.}
  
  \centering
  \begin{tabular}{lllllll}
    \multicolumn{2}{c}{}             \\
    Network   & Task & Parameters  & ROC AUC & Accuracy\\
    \hline
    \hline
    \multirow{3}*{WavPool} 
& MNIST & 182192 & \textbf{0.953$\pm$0.001} & \textbf{0.916$\pm$0.002} \\ 
& Fashion-MNIST & 182192 & 0.907$\pm$0.001 & 0.832$\pm$0.001 \\ 
& CIFAR & 235964 & \textbf{0.629$\pm$0.004} & \textbf{0.331$\pm$0.007} \\
    \hline
    \multirow{3}*{MLP Block} 
& MNIST & 209890 &  0.942 & 0.895$\pm$0.001 \\ 
& Fashion-MNIST & 209890 & \textbf{0.91$\pm$0.001} & \textbf{0.839$\pm$0.002} \\ 
& CIFAR & 273250& 0.604$\pm$0.002 & 0.288$\pm$0.003 \\
    \hline
    \multirow{3}*{CNN Block} 
& MNIST & \textbf{6806} & 0.95 & 0.91 \\ 
& Fashion-MNIST & \textbf{6806} & 0.898 & 0.815 \\ 
& CIFAR & \textbf{9046} & 0.603$\pm$0.003 & 0.285$\pm$0.006 \\

    \label{table:Network Comparison, Unoptimized}
  \end{tabular}
\end{table}

First, we perform these experiments with non-optimized parameters, for the purpose of showing how each block performs without an in-depth parameter exploration period. 
The parameters are chosen by leaving most parameters as the default parameters of PyTorch. 
The results of the non-optimized experiments can be found in Tab.~\ref{table:Network Comparison, Unoptimized}.
These values are the average of the scores obtained on the validation sets of the three different runs.
Values without error bars did not have significant variation between runs.

\begin{table}[b]
  \caption{The same networks and datasets from table \ref{table:Network Comparison, Unoptimized}, but optimized using a Bayesian Optimization scheme. 
  }
  
  \centering
  \begin{tabular}{lllllll}
    \multicolumn{2}{c}{} \\
    Network   & Task & Parameters  & ROC AUC & Accuracy\\
    \hline
    \hline
    \multirow{3}*{WavPool} 
& MNIST & 186722  & \textbf{0.978$\pm$0.001} & \textbf{0.96$\pm$0.001} \\ 
& Fashion-MNIST & 186722 & \textbf{0.925} & \textbf{0.865} \\ 
& CIFAR & 201672 & \textbf{0.658} & \textbf{0.386$\pm$0.001} \\
    \hline
    \multirow{3}*{MLP Block} 
& MNIST & 216250 & 0.947 & 0.905\\ 
& Fashion-MNIST & 217045 & 0.913$\pm$0.001 & 0.843$\pm$0.001 \\ 
& CIFAR & 215290 & 0.599 & 0.279 \\
    \hline
    \multirow{3}*{CNN Block} 
& MNIST & \textbf{96375} & 0.977$\pm$0.002 & 0.959$\pm$0.003 \\ 
& Fashion-MNIST & \textbf{154154} & 0.887$\pm$0.012 & 0.797$\pm$0.02 \\ 
& CIFAR-10 & \textbf{125175} & 0.642$\pm$0.01 & 0.352$\pm$0.019 \\ 

    \label{table:Network Comparison, Optimized}
  \end{tabular}
\end{table}

\begin{table}[t]
\caption{The fractional increase in monitored validation metrics of the WavPool block as compared to the other tested blocks. 
Results are shown for the optimized blocks.
The loss is minimized during training.
The improvement of the WavPool over other blocks is denoted by a negative sign for loss, but a positive sign for other metrics. 
WavPool provides a significant increase in accuracy without using a higher parameter count, as compared to the MLP.}

\centering 
    \begin{tabular}{lllll} 
    
        \multicolumn{2}{c}{} \\
         & Networks & Loss & ROC AUC & Accuracy \\ 
        
        \hline
        \hline
        \multirow{2}*{MNIST} 
& MLP & -0.478 & 0.03 & 0.056 \\ 
& CNN & -0.452 & 0.014 & 0.001 \\ 
        
        \hline
        \multirow{2}*{Fashion-MNIST} 
& MLP & -0.143 & 0.011 & 0.021 \\ 
& CNN & -0.333 & 0.039 & 0.085 \\ 
        
        \hline
        \multirow{2}*{CIFAR-10} 
& MLP & -0.123 & 0.098 & 0.416 \\ 
& CNN & -0.001 & 0.017 & 0.097 \\ 
        \label{table:gains} \\ 
        
    \end{tabular} \\ 
\end{table}

\begin{figure}[t]
    \centering
    \includegraphics[width=0.48\textwidth]{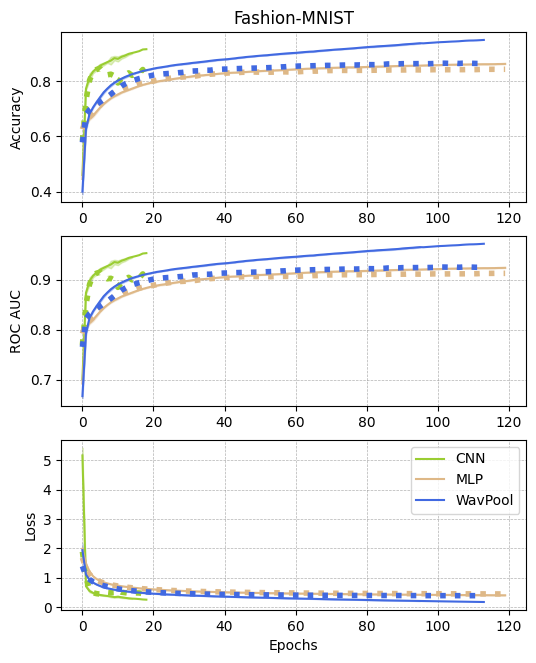}~~\includegraphics[width=0.490\textwidth]{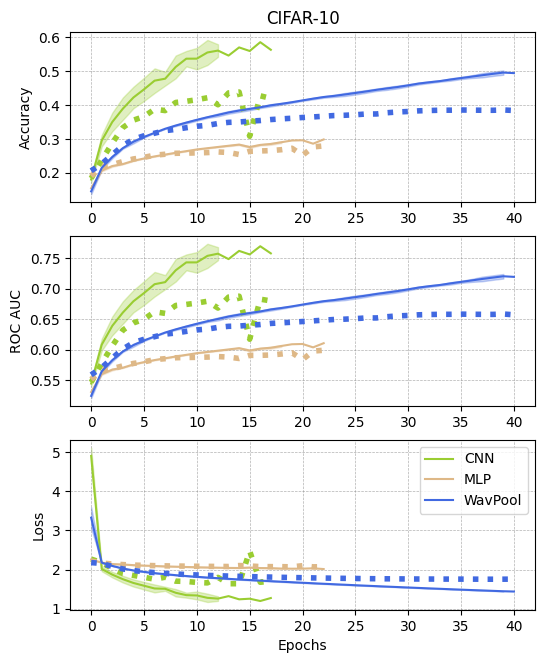} \\
    
    \caption{
        Performance of optimized networks on Fashion MNIST and CIFAR-10.
        The error bars show the variation of three trials with different network initialization and different dataset subsets. 
        Validation for a single trial is represented with a dashed line.
    }    
    \label{fig:lossacc_compare}
\end{figure}

Next, we use Bayesian hyperparameter optimization methods \cite{bayesopt} to find the best possible networks for each architecture and task combination. 
This was done three times with the goal of demonstrating the best capabilities of each network. 
This includes optimizing the learning rate, hidden layer sizes, and optimizer for all networks, and additionally the convolution kernel size and number of hidden channels for the CNN. 
The exact range of the optimization space can be found in App.~\ref{sec:appendix_space}. 
The quality of a network is assessed after an abbreviated training period of 20 epochs, based on the F1 score on the validation set.
The final network is then re-initialized and allowed to run for up to 120 epochs subject to early stopping as in the non-optimized case.
The results of the experiments on optimized networks are shown in Tab.~\ref{table:Network Comparison, Optimized}.
We also provide a relative comparison of performance on these metrics in Tab.~\ref{table:gains}.
We show the results of these experiments in Fig.~\ref{fig:lossacc_compare}.
The solid lines show the mean training performance, the shaded regions show the envelope of the performance across the three runs, and the dotted lines shown the results from the best-performing validation run.

WavPool requires significantly more trainable parameters than a single-channel CNN block with comparable architecture and complexity. 
This is attributed to the MicroWav's use of dense layers as means of holding trained component information.
Reduction in the size of these dense layers is still feasible, as no operations (e.g., weight-pruning or quantization) have been applied to these dense layers in the MicroWav blocks, nor have they been individually tuned for ideal size given the size of the wavelet transform output.
Because many signals are sparse in the wavelet representation \cite{10.5555/1525499}, further optimization of the WavPool network is an interesting and promising topic for future work.
Nevertheless, the results show that the WavPool is less information-hungry than the CNN.
For the MNIST and Fashion MNIST tasks, the number of parameters in the WavPool only increases by 2\% from the non-optimized to the optimized version, while the optimized versions of the CNN require many more parameters than the non-optimized version. 
For WavPool, the number of parameters does not significantly increase for the Fashion-MNIST and MNIST tasks.
The number of parameters decreases for CIFAR-10, while increasing predictive power significantly.

For both non-optimized and optimized experiments, the WavPool outperforms the MLP and the CNN with respect to accuracy and AUC ROC for almost all classification problems.
The exception is that the MLP outperforms WavPool at the sub-percent level on Fashion-MNIST in the non-optimized case. 
Despite its relatively high complexity (e.g., more independent layers) compared to a basic MLP, the WavPool contains at least $10^{4}$ fewer trainable parameters in all cases.
Trainable weights in the network are the ones in each MicroWav layer (see Fig.~\ref{fig:microwav}) and on the final dense layer of the WavPool (see Fig.~\ref{fig:wavpool_arch}).
We show this from the increase in accuracy across the different blocks tested.
While the blocks were exposed to the same amount of data in identical training schemes, the wavelet showed a $\sim10\%$ increase in accuracy over a standard MLP and a CNN block for tests on CIFAR-10. 
CIFAR-10 is itself a real image dataset, so the correlation of the pixels within a given image is both local and non-local \cite{Krizhevsky2009LearningML}. 
The decomposition of the input applied by WavPool is lossless, yet contains non-local and local information broken into different views of the data.
The initial decomposition allows different parts of the network to learn information independently.

\section{Conclusion}

In this work, we demonstrate the functionality and efficacy of a multi-resolution decomposition (MRD) within a neural network. 
Specifically, we decompose data using a Haar wavelet and train classifiers on this stacked, decomposed data.
On each level of the transformed data, we apply a dense layer, which we refer to as a MicroWav unit.
By training all of the MicroWav units in parallel, we create a multi-resolution perceptron (MRP), on which we use pooling to create the WavPool block.
We observe better performance with this classifier relative to a comparable CNN on MNIST, Fashion-MNIST, and CIFAR-10 datasets.

It is notable that WavPool has the capacity to outperform a CNN with comparable architectural size and complexity.
CNNs abstract scale information from an image using fixed filter sizes and dynamic, trainable weights.
The abstraction comes at the expense of increasing the size of the data vector in some implementation scenarios.
In contrast, the MRP uses filters with fixed values, but varying in kernel size/spatial extent, to partition the data into components of varying size without increasing the size of the input data vector.
We conclude that the multi-resolution aspect of the MRP makes information from all scales of the image readily available to the classifier.
In addition, on multichannel data, we expect that WavPool will have better scaling properties while retaining its high performance.

The MRP also holds promise for a wider range of applications.
Similar techniques in combination with CNNs have shown promising, near-state-of-the-art performance on a variety of tasks \cite{2017arXiv171207493C, 2017arXiv170403915L, DBLP:journals/corr/abs-1805-07071, DBLP:journals/corr/abs-1805-08620, 2022arXiv220704978Y}.
Supplementing CNNs with MRPs or MicroWav-like blocks in these architectures is a promising route for future investigation.
There is also potential for this method in Quantum Neural Network contexts. 
For example, the Haar wavelet is an operation requiring only 1 or -1, (as seen in equation \ref{eq:Haar1D} and \ref{eq:Haar2D}), so it can be easily implemented in the schema required for quantum neural networks, because the operations involved in the Haar wavelet mimic those used on qubits \cite{kwak2021quantum}.

\section*{Acknowledgments}
We acknowledge the Deep Skies Lab as a community of multi-domain experts and collaborators who've facilitated an environment of open discussion, idea-generation, and collaboration. This community was important for the development of this project.
We thank Aleksandra \'Ciprijanovi\'c, Andrew Hearin, and Shubhendu Trivedi for comments on the manuscript.
This manuscript has been authored by Fermi Research Alliance, LLC under Contract No.~DE-AC02-07CH11359 with the U.S.~Department of Energy, Office of Science, Office of High Energy Physics.

\bibliography{thebib}

\newpage
\appendix

\begin{table}[h]
\caption{A glossary of terms used in this work.}
\centering
  \begin{tabular}{m{3.9cm}llm{5.9cm}}
    Term & Symbol & Ref.  & Definition \\
    \hline
    signal & $S$ & - & input of arbitrary size and dimension \\ \hline
    view & - & - & transform of an input \\ \hline
    multi-resolution analysis & - & - & view of $S$ incorporating multiple length scales \\ \hline
    wavelet & - & - & integrable function with compact support \\ \hline
    smoothing wavelet & $\phi$ & - & wavelet for capturing local averages of $S$ \\ \hline
    differencing wavelet & $\psi$ & - & wavelet for capturing local variations of $S$\\ \hline
    Daubechies-$n_v$ wavelet & - & - & wavelet with $n_v$ vanishing moments \\ \hline
    Haar wavelet & - & Eqs.~(\ref{eq:Haar1D}, \ref{eq:Haar2D}) & synonym for Daubechies-1 wavelet \\ \hline
    smooth views & $C_\ell$ & Eq.~\eqref{eq:Ci} & local, scale-dependent averages of $S$ \\ \hline
    details & $W_\ell$ & Eq.~\eqref{eq:Wi} & local, scale-dependent variations of $S$ \\ \hline
    level & $\ell$ & - & identifier for the {\it resolution} of the detail, determined by the number of times $S$ has been smoothed \\ \hline
    number of levels & $L$ & Eq.~\eqref{eq:num_levels} & maximum number of times $S$ can be smoothed \\ \hline
    dimension & D & - & dimensionality of $S$ (e.g., 2 for an image) \\ \hline
    multi-resolution decomposition & MRD, $\cM_S$ & Eq.~\eqref{eq:decomp} & lossless, non-multiplexed set of views of $S$, consisting of $(2^{\rm D}-1) \times L$ details and one smooth version of $S$\\ \hline
    MicroWav & - & Fig.~\ref{fig:microwav} & dense layers applied in parallel to the details of a single level of the MRD \\ \hline
    multi-resolution perceptron & MRP & Fig.~\ref{fig:wavpool_arch} & stack of MicroWavs \\ \hline
    WavPool & - & Fig.~\ref{fig:wavpool_arch} & padded, pooled, and trained MRP
    \end{tabular}
    \label{tab:glossary}
\end{table}
\section{Terminology}
\label{sec:App_Terminology}

In the interest of clarity, we provide in Tab.~\ref{tab:glossary} a glossary of terms used in this work.

We also remark on alternate terminology sometimes used in similar contexts.
Procedures related to the multi-resolution decomposition (MRD) are variously referred to as the ``multi-resolution analysis'', the ``wavelet transform'', or the ``wavelet decomposition.''
A multi-resolution analysis does not necessarily contain all of the information from the original data \cite{2011arXiv1101.2286M}, and a wavelet transform does not necessarily contain multiple scales \cite{daubechiesbook}.
Therefore, those are general terms that generate different views of the data.
In contrast, MRD refers specifically to a multi-scale wavelet transform that partitions the original data into a new set that is exactly the same size as the input.
There is a one-to-one correspondence between a signal and its MRD.

The smoothing and differencing wavelets are sometimes referred to by other names.
These are sometimes respectively called a {\it father wavelet} or {\it scaling function} and a {\it mother wavelet} or {\it wavelet function} in the literature. 
The terms we choose are selected for maximal clarity.

\section{Worked Examples}
\subsection{Analytic Decomposition of a $2\times2$ Matrix}
\label{sec:App_matrixdecomp}

In this appendix, we perform the multiresolution decomposition of a $2\times2$ matrix using the Haar wavelet. 
We also demonstrate the inversion procedure.

Let the input be general:
\begin{equation}
    A = \left( \begin{array}{cc} a & b \\ c & d \end{array} \right).
\end{equation}
Following the notation of Sec.~\ref{sec:method}, we have
\begin{equation}
    \begin{split}
    \phi_{1,2} = \frac12 \left( \begin{array}{cc} 1 & 1 \\ 1 & 1 \end{array} \right),& \qquad \psi_{1,2}^v = \frac12 \left( \begin{array}{cc} 1 & 1 \\ -1 & -1 \end{array} \right),\\
    \psi_{1,2}^h = \frac12 \left( \begin{array}{cc} 1 & -1 \\ 1 & -1 \end{array} \right),&\qquad  \psi_{1,2}^d = \frac12 \left( \begin{array}{cc} 1 & -1 \\ -1 & 1 \end{array} \right).
    \end{split}
\end{equation}
Then
\begin{equation}
    C = \phi_{1,2} \circ A, \qquad W_1^a = \psi_{1,2}^a \circ A
\end{equation}
or
\begin{equation}
    C = \frac{a+b+c+d}2,~~ W_1^v = \frac{a+b-c-d}2,~~ W_1^h = \frac{a-b+c-d}2,~~ W_1^d = \frac{a-b-c+d}2.
\end{equation}
The set $\{ C, W_1^v, W_1^h, W_1^d\}$ is four numbers, which is the same amount of data required to specify the original matrix $A$.

To reconstruct $A$, we must invert the MRD. 
The original data is obtained from the new data via
\begin{equation}
\begin{split}
    a = \frac{C + W_1^v + W_1^h + W_1^d}2, \qquad b = \frac{C + W_1^v - W_1^h - W_1^d}2,
    \\ c = \frac{C - W_1^v + W_1^h - W_1^d}2, \qquad d = \frac{C - W_1^v - W_1^h + W_1^d}2.
\end{split}
\end{equation}
Using matrix notation, we can expand the components of the MRD and recombine them to obtain $A$ directly:
\begin{equation}
    A = C \phi_{1,2} + \sum_{a=v,h,d} W_1^a \psi_{1,2}^a ,
\end{equation}
where all multiplications are re-scalings of the original wavelets.

\subsection{Worked Example: Image Decomposition}
\label{sec:App_imagedecomp}

\begin{figure}[t]
    \centering
    \includegraphics[width=\textwidth]{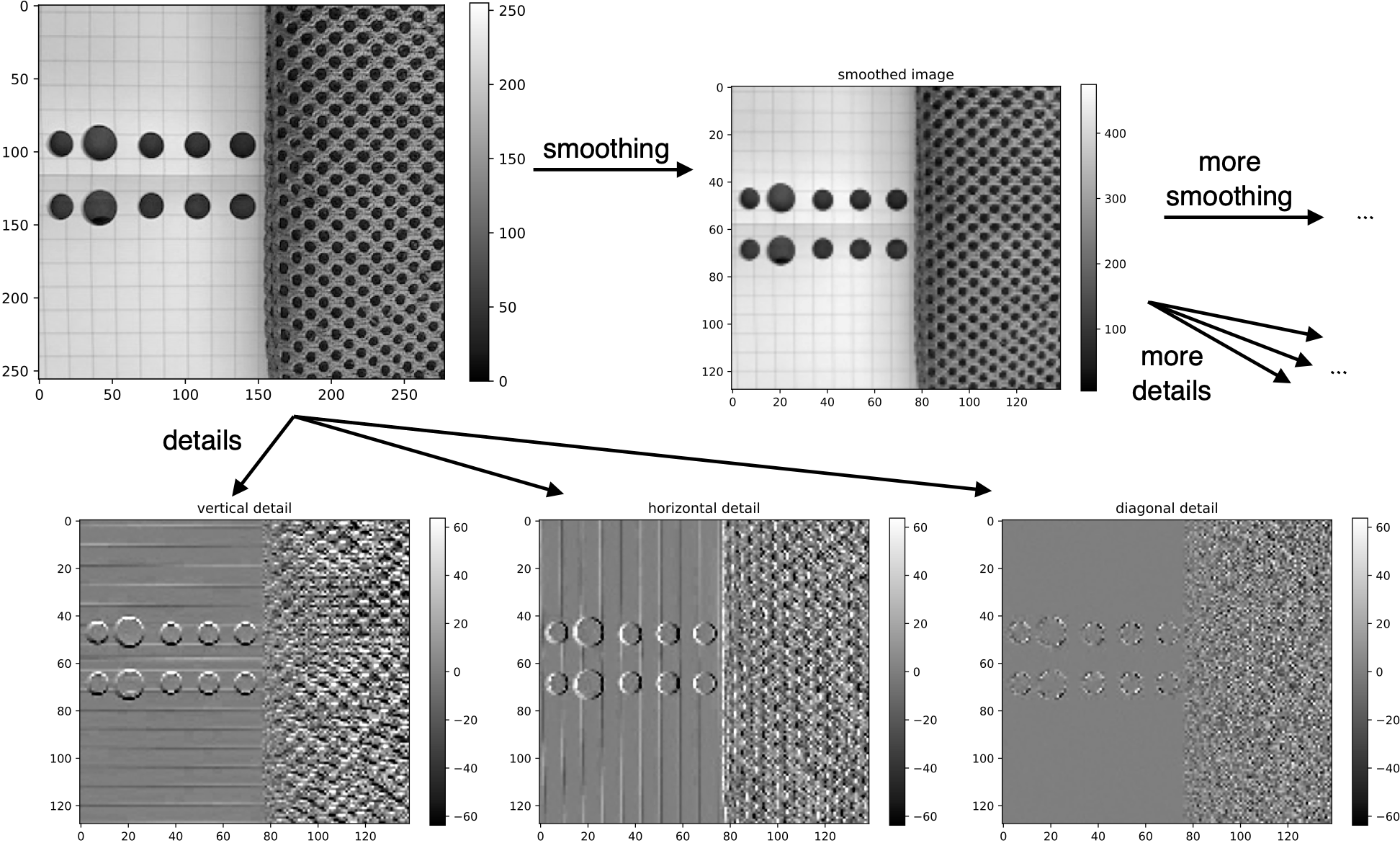}
    \caption{
        Example of the first level of an MRD.
        The image (top left) is convolved with the vertical, horizontal, and diagonal differencing wavelets as defined in Eq.~\eqref{eq:Haar2D}, and the resulting details are shown in the bottom row.
        The image can also be smoothed (top center), after which the vertical, horizontal, and diagonal differencing procedures can be iterated.
    }    
    \label{fig:mrd_cartoon}
\end{figure}

In Fig.~\ref{fig:mrd_cartoon} we show an example of an image that has been decomposed with a Haar wavelet.
We show only the first level of the MRD.
The input signal $S$ (top left) is a photograph that contains a four distinct textures: circles, vertical lines, horizontal lines, and a slightly irregular web of shapes.
The vertical detail (bottom left) reveals the presence of noticeable variations across $S$ as compared vertically between neighboring rows of the image.
The horizontal detail (bottom middle) reveals the presence of noticeable variations across $S$ as compared horizontally between neighboring columns of the image.
The diagonal detail (bottom right) reveals the presence of noticeable variations across $S$ as compared diagonally between neighboring patches of the image.
The left half of the image has vertical and horizontal textures at the locations of the grid lines and near the circles, but only has evident diagonal structure near the edges of the circles.
The right half of the image has structure in all directions.

The smoothed image (top center) is a lower-resolution version of $S$.
It will be used as the input for the next layer of the MRD.
Each of the three details and the smoothed image contain one-quarter as many pixels as $S$ does, demonstrating that the size of the data vector has not increased at this level of the decomposition.
With each additional level of the transform, each detail and smoothed version will be one-quarter the size of the smoothed image at the preceding level.
This process will continue until smoothing is no longer possible, which occurs at level $L$.

\section{Additional Comparisons}
\subsection{Comparison of network timing}

While measuring the performance of the multiple training configurations, the number of network parameters, single datum inference time, and training time were also recorded. 
Training time varies largely due to the early stopping applied to the training loop. 

\begin{figure}[t]
    \centering
        \includegraphics[width=0.3\textwidth]{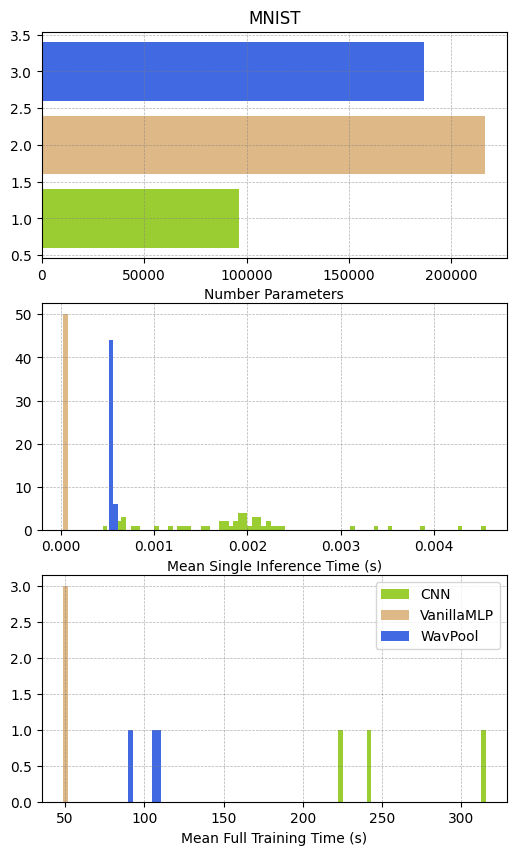}~~\includegraphics[width=0.31\textwidth]{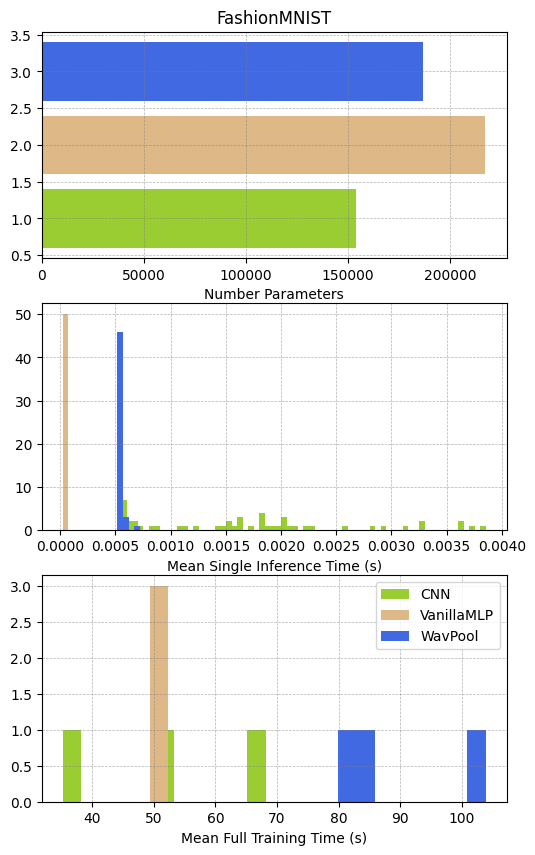} ~~ 
        \includegraphics[width=0.3\textwidth]{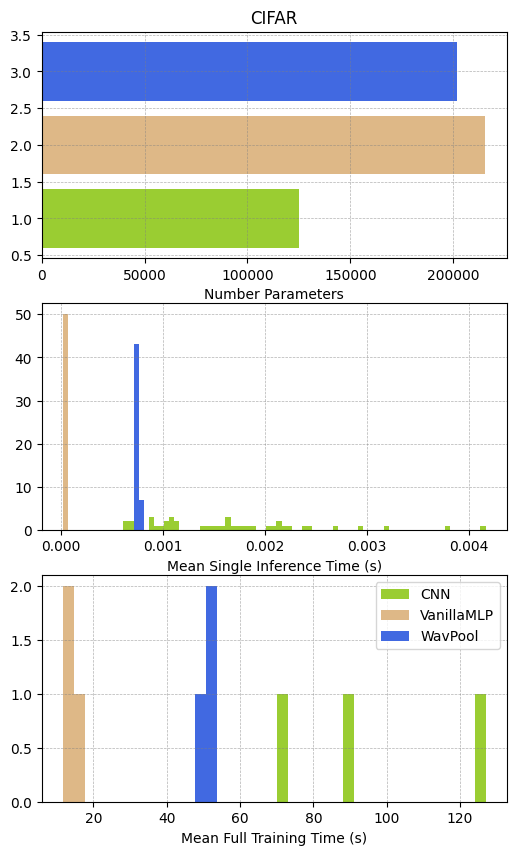} \\

    \caption{Timing and parameter count comparison of the optimized networks on MNIST, Fashion-MNIST, and CIFAR-10. The differences in timing between the WavPool and MLP can be attributed to the pure-python backend for the wavelet transforms.}
    \label{fig:mnist_params}
\end{figure}

\subsection{Confusion Matrices}

\begin{figure}[t]
    \centering
    \includegraphics[width=0.85\textwidth]{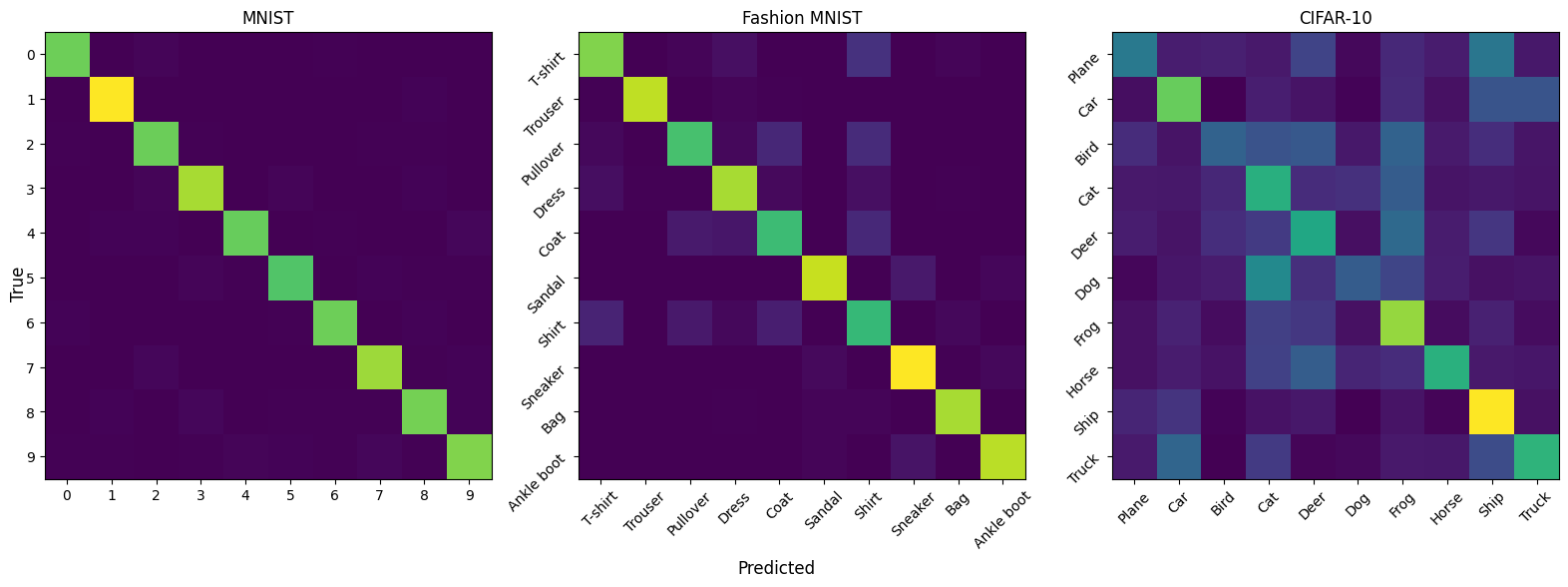} ~~ 
    \includegraphics[width=0.85\textwidth]{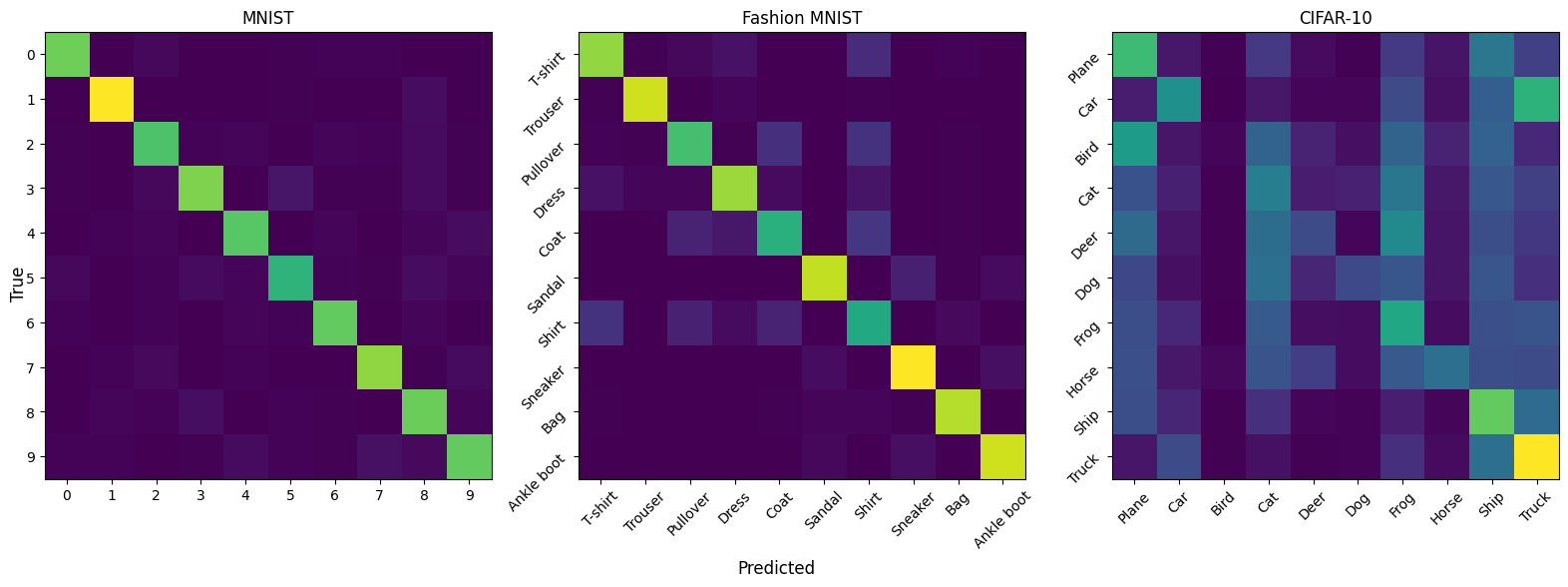} ~~ 
    \includegraphics[width=0.85\textwidth]{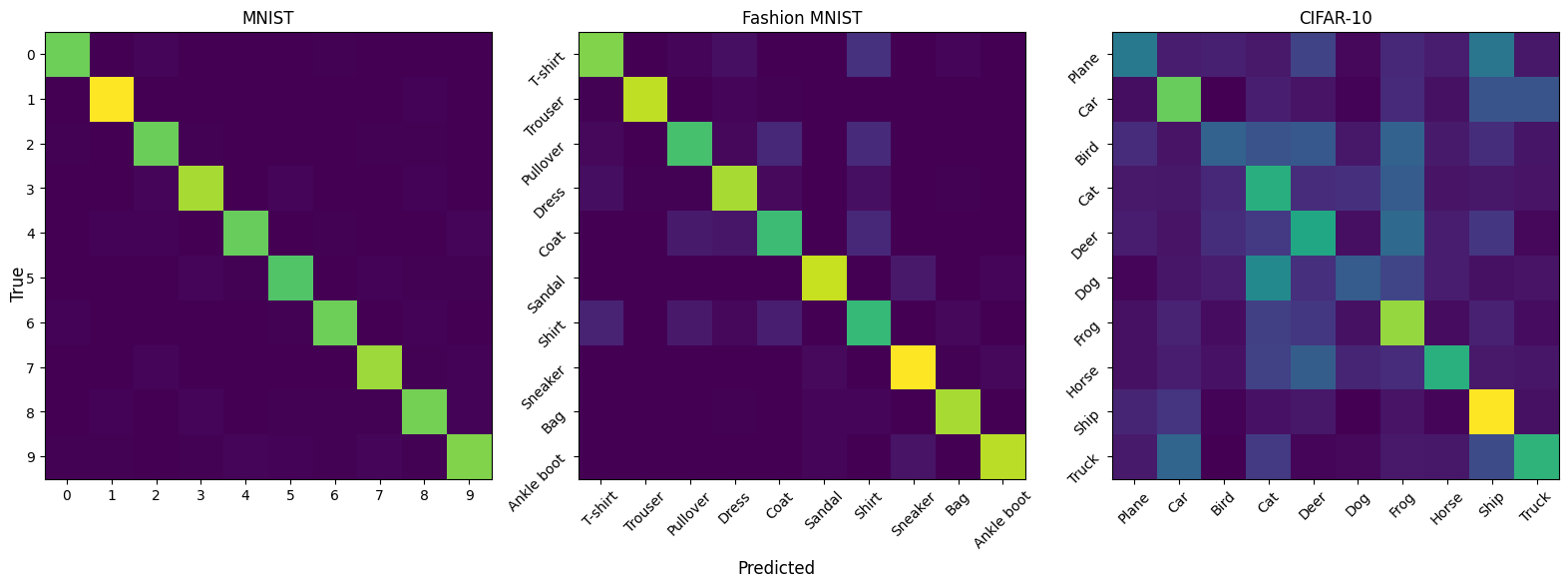} \\
    
    \caption{
    Confusion matrices for results on an independent test set for each of the three outlined tasks. 
    WavPool (Top). 
    CNN (Middle). 
    MLP (Bottom). 
    }    

    \label{fig:confusion}
\end{figure}

In Fig.~\ref{fig:confusion} we show the confusion matrices for the three different networks.

\section{Parameter space explored during optimization}
\label{sec:appendix_space}

During our optimization experiments, spaces from which to select parameters were constructed. These spaces are outlined in tables \ref{tab:Wav_param_space}, \ref{tab:MLP_param_space}, and \ref{tab:CNN_param_space}. 

\begin{table}[t]
    \centering
    \begin{tabular}{c|c|c|c|c}
         & Learning Rate & Hidden Size & Pooling Size & Hidden Layer Scaling \\
         Minimum & $10^{-6}$ & 200 & 2 & true\\ 
         Maximum & 0.8 & 300 & 4 & false 
    \end{tabular}
    \caption{Parameter Space explored for the WavPool during optimization}
    \label{tab:Wav_param_space}
\end{table}

\begin{table}[h]
    \centering
    \begin{tabular}{c|c|c}
         & Learning Rate & Hidden Size \\
         Minimum & $10^{-6}$ & 200\\ 
         Maximum & 0.8 & 300
    \end{tabular}
    \caption{Parameter Space explored for the MLP during optimization}
    \label{tab:MLP_param_space}
\end{table}

\begin{table}[h]
    \centering
    \begin{tabular}{c|c|c|c|c}
         & Learning Rate & Kernel Size & Hidden Channels$_1$ & Hidden Channels$_2$ \\
         Minimum & $ 10^{-6}$ & 2 & 1 & 1\\ 
         Maximum & $0.8$ & 4 & 20 & 20 \\
    \end{tabular}
    \caption{Parameter Space explored for the CNN during optimization}
    \label{tab:CNN_param_space}
\end{table}

\section{Limitations}
\label{sec:app_limitations}

For all results show in our work, we used the Daubechies-1, or Haar, wavelet.
This wavelet is known to have suboptimal frequency response characteristics \citep{Dremin:2001kv}.
Using any other wavelet or signal that provides a complete orthonormal basis of D-dimensional inputs, including the Daubechies-$n_v$ wavelets for $n_v>1$, should offer similar results as long as all scales of the MRD are utilized, but this deserves future quantitative study.

We also limit our study to single-channel images, and while we believe our results can be extended to higher dimensional tasks, this is not shown in our experimentation.

\end{document}